\RequirePackage{snapshot}
\documentclass[]{article}
\pdfoutput=1

\usepackage[utf8]{inputenc}
\usepackage{graphicx}
\graphicspath{{figures/}}

\usepackage[style=authoryear,sorting=anyt,isbn=false,backend=bibtex,doi=false,maxcitenames=2]{biblatex}
\usepackage{tikz}
\usepackage{tkz-euclide}

\newcommand\noktikz{}

\title{DeepGaze II: Reading fixations from deep features trained on object recognition}
\author{Matthias Kümmerer \and Thomas S. A. Wallis \and Matthias Bethge}

\begin{document}
\maketitle
\begin{abstract}
	Here we present DeepGaze II, a model that predicts where people look in images.
	The model uses the features from the VGG-19 deep neural network trained to identify objects in images.
	Contrary to other saliency models that use deep features, here we use the VGG features for saliency 
	prediction with no additional fine-tuning (rather, a few readout layers are trained
	on top of the VGG features to predict saliency).
	The model is therefore a strong test of transfer learning.
	After conservative cross-validation, DeepGaze II explains about 87\% of the explainable information gain in the patterns of
	fixations and achieves top performance in area under the curve metrics on the
	MIT300 hold-out benchmark.
	These results corroborate the finding from DeepGaze I (which explained 56\% of the explainable information gain),
	that deep features trained on object recognition provide a versatile feature space
	for performing related visual tasks. 
	We explore the factors that contribute to this success and present several informative image examples.
	A web service is available to compute model predictions at \url{http://deepgaze.bethgelab.org}.
\end{abstract}

\section{Introduction}

% why care about saliency? what is it?
Humans and other animals with foveated visual systems make several eye movements per second, bringing their high-resolution 
fovea to bear on things they want to see.
Understanding the factors that guide eye movements is therefore an important component of understanding behaviour.
One problem that has received significant attention is that of predicting fixation locations given the image the observer is viewing (usually in a free-viewing paradigm).
Here we term this problem \textit{saliency prediction}, in keeping with the computer vision literature \footnote{Note that \textit{saliency} is sometimes defined as the visibility or contrast of some image region, irrespective of whether that predicts human fixations.}.

% deep features and deep gaze I.
The state-of-the-art in saliency prediction improved markedly since 2014 with the advent of models using deep neural networks. 
The first of these models \parencite{Vig2014} trained deep neural networks on the task of saliency prediction. 
We subsequently boosted performance significantly above EDN in our model DeepGaze I \parencite{kuemmerer2015deepgaze}, by using pretrained features (taken from AlexNet \cite{krizhevsky2012}) trained on the ImageNet object recognition benchmark.
This is therefore an example of transfer learning, where features learned on one task are re-used for a second task (with or without fine-tuning).
The success of this approach is exciting because it implies that the features learned by deep neural networks on ImageNet have abstracted generalisable information from images.
The transfer learning paradigm seems to be particularly important for saliency prediction because typical saliency datasets are relatively small---a few thousand images with fixations in the hundreds per image---making learning of deep neural networks from scratch \parencite{Vig2014} relatively unconstrained.

% Developments since DGI
Since DeepGaze I, a variety of new models also apply transfer learning approaches using deep features.
In contrast to DeepGaze I, which uses AlexNet, the SALICON model \parencite{Huang_2015_ICCV}, DeepFix \parencite{Kruthiventi2015} and PDP \parencite{jetley2016} use the better-performing VGG-19 network \parencite{Simonyan2014}, whose features are retrained on saliency prediction using the SALICON dataset then fine-tuned on the MIT1003 dataset.
SALICON and DeepFix substantially improved performance over DeepGaze I in the MIT benchmark (\cite{mit-saliency-benchmark}; see below).
The scale of this improvement could suggest that retraining deep features is crucial for further performance improvement, or it could suggest that the VGG features themselves (which significantly outperform AlexNet for object recognition) provide a better feature space for saliency prediction irrespective of retraining.
In this paper we show the latter is the case.

% Deep Gaze II
Here, we introduce DeepGaze II.
Relative to DeepGaze I, it uses the VGG-19 pretrained network and pretraining on the SALICON dataset.
In addition, rather than using a linear predictor, DeepGaze II uses a pointwise nonlinear combination of deep features.
Two additional crucial distinctions between DeepGaze II and the models discussed above (SALICON, DeepFix and PDP) are that we train our  model in a probabilistic framework optimising the log-likelihood \parencite{kuemmerer2015}, and that we do not re-train the VGG features themselves.
DeepGaze II (as for DeepGaze I) also models the centre bias as an explicit prior.

\section{Methods}

\subsection{Model}

% note changes from DeepGaze I.
\begin{figure}
  \begin{center}
    \input{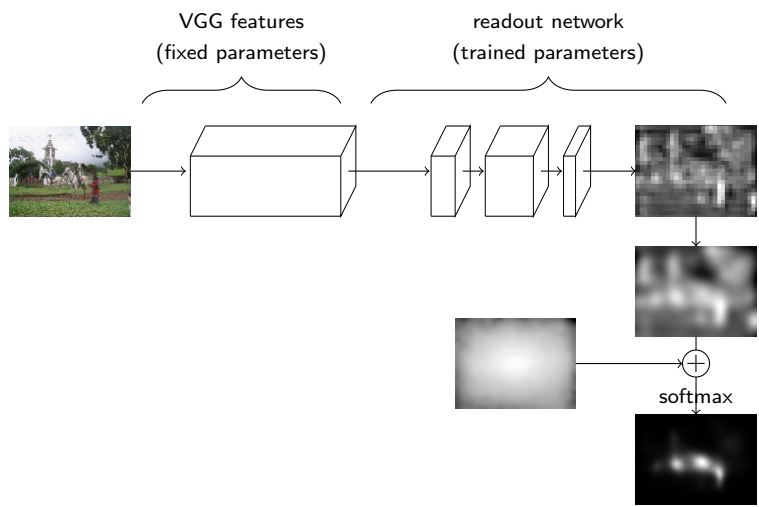}
  \end{center}
  \label{fig:architecture}
  \caption{The architecture of DeepGaze II. 
	The activations of a subset of the VGG feature maps for a given image are passed to a second neural network (the readout network) consisting 
	of four layers of $1 \times 1$ convolutions.
	The parameters of VGG are held fixed through training (only the readout network learns about saliency prediction). 
	This results in a final saliency map, which is then blurred, combined with a centre bias and converted into a probability 
	distribution by means of a softmax.
  }
\end{figure}

% formulation as probabilistic model.
As for DeepGaze I, we formulate DeepGaze II as a probabilistic model.
%If we have multiple models that provide predictions of fixation locations from images, how should we compare them?
%Several benchmarking platforms exist, including the MIT saliency benchmark, SALICON and LSUN.
%These allow the submission of model predictions that are then evaluated on a held-out test set, providing a measure of how well the models can predict data on which they were never trained.
%These platforms compare models on numerous evaluation criteria that can produce inconsistent model rankings.
Building on previous work applying probabilistic modelling to fixation prediction \parencite{vincent2009, Barthelme}, we have recently shown that formulating existing models appropriately can remove most of the inconsistencies between existing model evaluation metrics \parencite{kuemmerer2015}. % cite geisler, when he published.
Furthermore, we argued that using log-likelihood (the standard way to compare probabilistic models) as an evaluation criteria represented a useful and intuitive loss function for model evaluation with close ties to information theory (though other loss functions may have advantages for certain use cases \parencite{Vig2014}.
Here, we train and evaluate DeepGaze II using the framework of log-likelihood (specifically reported as information gain explained, see \cite{kuemmerer2015}) for our in-house tests, and present key metrics from the MIT benchmark (AUC, shuffled AUC).

%%% Architecture

The architecture of DeepGaze II is visualized in \ref{fig:architecture}. 
The image in question is (possibly after resizing, see below) given as input to the VGG-19 network, from which all fully-connected layers have been removed and for which all filters have been rescaled to yield feature maps with unit variance over the imagenet dataset \parencite{gatys2015}.
After processing the image in VGG, the feature maps of a selection of layers (conv5\_1, relu5\_1, relu5\_2, conv5\_3, relu5\_4; selected via random search) are rescaled and cropped to match an earlier layer (conv2\_1 in our implementation).
This rescaling is necessary to equate the sizes of the feature maps from different layers; conv2\_1 is chosen such that spatial resolution is sufficient for precise prediction but computation time is reduced. 
Matching here means that we identify a pixel in the output of a convolution with the center of its receptive field in its input layer.
%We find this more appealing than simply resizing because resizing could lead to 

After rescaling and cropping, these feature maps have the same size and can be combined into one 3-dimensional tensor (with $5 \times 512$ channels) which is used as input for a second neural network (called the \textit{readout network}) in the following.
This readout network consists of four layers of 1x1 convolutions followed by ReLu nonlinearities.
Therefore, the readout network is only able to represent a \textit{pointwise} nonlinearity in the VGG features.
The first three layers use 16, 32, and 2 features.
The last layer has only one output channel $O(x, y)$.
This final output from the readout network is convolved with a Gaussian to regularize the predictions:
\[
  S(x, y) = O(x, y) \star G_\sigma
\]

Fixations tend to be near to the center of the image in a way which is strongly task and dataset dependent \parencite{Tatler2007Centerbias}.
Therefore it is important to model this center bias and do so in a way that allows easy substitution of other centre biases (e.g. depending on the task).
We do so by explicitly modelling the center bias as a prior distribution that is added to $S$:
\[
  S'(x, y) = S(x, y) + \log p_\mathrm{baseline}(x, y)
\]
$S'(x, y)$ is finally converted into a probability distribution over the image by the means of a softmax (as for DeepGaze I):
\[
  p(x, y) = \frac{\exp(S'(x, y))}{\sum_{x, y} \exp(S'(x, y))}
\]

In implementing DeepGaze II, we use a caffe \parencite{jia2014caffe} implementation for the VGG; all other parts of the model are implemented in Lasagne and Theano \parencite{theano2016}.

\subsection{Training}

% include new figure describing crossval.
\begin{figure}
\begin{center}
  \usetikzlibrary{calc}
\usetikzlibrary{shapes}

\newcommand\scalingfactor{0.5}
\begin{tikzpicture}[scale=\scalingfactor,font=\sffamily\small,every node/.style={inner sep=0,outer sep=0},
                     box/.style={line width=0.01cm},
                     label/.style={font=\footnotesize\sffamily},
                     trainnode/.style={box},
                     trainlabel/.style={label},
                     testnode/.style={box,color=red,fill=red!20!white},
                     testlabel/.style={label,color=red},
                     modelconnection/.style={->,line width=0.01cm, dashed},
                     headline/.style={font=\large\sffamily}]

  %%%%%%%%%%%%%%%%%%%%%%%%%%%%%%%%%%%%%%%
  %%     Coordinate Setup
  %%%%%%%%%%%%%%%%%%%%%%%%%%%%%%%%%%%%%%%

  \coordinate (headlines_vpos) at (0,6);

  \coordinate (dataset_height) at (0, 1);
  \coordinate (salicon_width) at (6,0);
  \coordinate (mit_pretraining_width) at (1.9,0);
  \coordinate (line_width) at (0.1cm,0);
  \coordinate (part_sep) at (2.5, 0);

  \coordinate (vert_sep) at (0, 1);
  \coordinate (finetuning_mit_width) at (6,0);
  \coordinate (evaluation_mit300_width) at (1.5, 0);

  \coordinate (dots_sep) at (0, 0.5);

  \coordinate (legend) at (0, 4);
  \coordinate (legend_vsep) at (0,-1.3);
  \coordinate (legend_box_width) at (1,0);
  \coordinate (legend_hsep) at (1.5, 0);

  %% Derived coordinates

  \coordinate (vertical_center) at ($ 0.5*(dataset_height) $);
  \coordinate (dataset_height_half) at ($ 0.5*(dataset_height) $);
  \coordinate (total_row_height) at ($ (dataset_height) + (vert_sep) $);
  \coordinate (salicon_center) at ($ 0.5*(salicon_width) + (dataset_height_half) $);
  \coordinate (mit_pretraining) at (salicon_width);
  \coordinate (mit_pretraining_center) at ($ (mit_pretraining) + 0.5*(mit_pretraining_width) + (dataset_height_half) $);
  \coordinate (pretraining_right_end) at ($ (salicon_width) + (mit_pretraining_width) + (dataset_height_half) $);
  \coordinate (pretraining_headline) at ($ 0.5*(salicon_width) + 0.5*(mit_pretraining_width) + (headlines_vpos) $);

  \coordinate (finetuning_part) at ($ (salicon_width) + (mit_pretraining_width) + (part_sep) $);
  \coordinate (finetuning_mit_test_width) at ($ 0.1*(finetuning_mit_width) $);
  \coordinate (finetuning_center) at ($ (vertical_center) + (finetuning_part) + 0.5*(finetuning_mit_width) - 0.5*(total_row_height) -0.3*(vert_sep) + (dataset_height_half) $);
  \coordinate (finetuning_headline) at ($ (finetuning_part) + 0.5*(finetuning_mit_width) + (headlines_vpos) $);

  \coordinate (evaluation_part) at ($ (finetuning_part) + (finetuning_mit_width) + (part_sep) $);
  \coordinate (evaluation_center) at ($ (evaluation_part) + 0.5*(evaluation_mit300_width) + (dataset_height_half) $);
  \coordinate (evaluation_left_end) at ($ (evaluation_part) + (dataset_height_half) $);
  \coordinate (evaluation_headline) at ($ (evaluation_part) + 0.5*(evaluation_mit300_width) + (headlines_vpos) $);

  %%%%%%%%%%%%%%%%%%%%%%%%%%%%%%%%%%%%%
  %%       Actual picture
  %%%%%%%%%%%%%%%%%%%%%%%%%%%%%%%%%%%%%

  %% Headlines

  \node[headline] at (pretraining_headline) {Pretraining};
  \node[headline] at (finetuning_headline) {Fine-tuning};
  \node[headline] at (evaluation_headline) {Evaluation};

  %% Pretraining

  \draw[trainnode] (0,0) rectangle + ($ (dataset_height) + (salicon_width) $);
  \node[trainlabel] at (salicon_center) {SALICON};
  \draw[testnode] (salicon_width) rectangle + ($ (dataset_height) + (mit_pretraining_width) - (line_width) $);
  \node[testlabel] at (mit_pretraining_center) {\tiny{MIT1003}};

  %% Finetuning

  % Heading Box

  \coordinate (this_location) at ($ (vertical_center) + (finetuning_part) + 1.5*(total_row_height) - 0.5*(vert_sep) $);
  %\draw[trainnode] (this_location) rectangle + ($ (finetuning_mit_width) + (dataset_height) $);
  \node[trainlabel] at ($ (this_location) + 0.5*(finetuning_mit_width) + (dataset_height_half) $) {MIT1003};

  % Train/Test Boxes

  \foreach \x in {1,2} {
    \coordinate (this_location) at ($ (vertical_center) + (finetuning_part) + 2.5*(total_row_height) -0.5*(vert_sep) - \x*(total_row_height) $);
    \draw[trainnode] (this_location) rectangle + ($ (finetuning_mit_width) + (dataset_height) $);

    \coordinate (this_test_location) at ($ (this_location) + (finetuning_mit_width) - \x*(finetuning_mit_test_width) $);
    \draw[testnode] (this_test_location) rectangle + ($ (finetuning_mit_test_width) + (dataset_height) $);

    \draw[modelconnection] (pretraining_right_end)-- ($ (this_location) + (dataset_height_half) $);
    \draw[modelconnection] ($ (this_location) + (finetuning_mit_width) + (dataset_height_half) $) -- (evaluation_left_end);
  }

  \node at (finetuning_center) {10 folds};
  \node at ($ (finetuning_center) + 1.3*(dots_sep) $) {$\vdots$};
  \node at ($ (finetuning_center) - (dots_sep) $) {$\vdots$};

  \coordinate (this_location) at ($ (vertical_center) + (finetuning_part) - 1.5*(total_row_height) $);
  \draw[trainnode] (this_location) rectangle + ($ (finetuning_mit_width) + (dataset_height) $);

  \coordinate (this_test_location) at ($ (this_location) $);
  \draw[testnode] (this_test_location) rectangle + ($ (finetuning_mit_test_width) + (dataset_height) $);
  \draw[modelconnection] (pretraining_right_end)-- ($ (this_location) + (dataset_height_half) $);
  \draw[modelconnection] ($ (this_location) + (finetuning_mit_width) + (dataset_height_half) $) -- (evaluation_left_end);

  %% Evaluation

  \draw[testnode] (evaluation_part) rectangle + ($ (dataset_height) + (evaluation_mit300_width) $);
  \node[testlabel] at (evaluation_center) {\tiny{MIT300}};

  %% Legend

  \coordinate (this_location) at (legend);
  \draw[trainnode] (this_location) rectangle + ($ (legend_box_width) + (dataset_height) $);
  \node[anchor=west] at ($ (this_location) + (legend_hsep) + (dataset_height_half) $) {training};

  \coordinate (this_location) at ($ (this_location) + (legend_vsep) $);
  \draw[testnode] (this_location) rectangle + ($ (legend_box_width) + (dataset_height) $);
  \node[anchor=west] at ($ (this_location) + (legend_hsep) + (dataset_height_half) $) {stopping criterion/evaluation};

\end{tikzpicture}
  \label{fig:crossvalidation}
  \caption{Training and crossvalidation procedure of the readout network used for DeepGaze II.
  In the pretraining phase, the model is trained on the 10000 images of the SALICON dataset
  using the 1003 images from the MIT1003 as a stopping criterion.
  In the fine-tuning phase, ten models are trained (starting from the pretrained model), each on 
  90\% of the MIT1003 data for training and a unique 10\% for stopping (10-fold crossvalidation).
  In our evaluation (reported below), for each image we use the model predictions from the model that
  did not use that image in training.
  The final model evaluation is performed via the MIT benchmark on the held-out MIT300 dataset, based 
  on a mixture of the ten models from the fine-tuning stage.
  }
  
\end{center}
\end{figure}
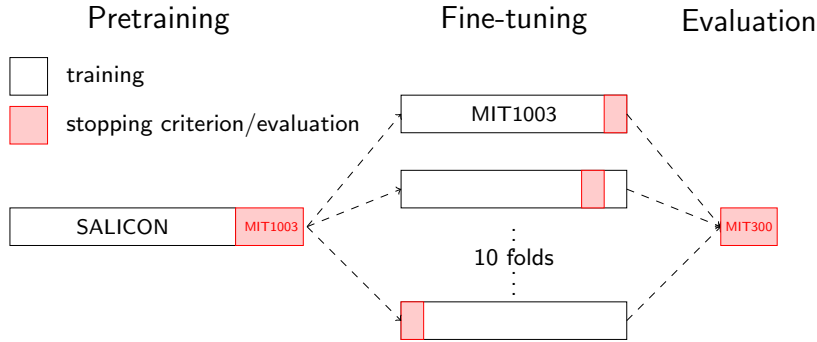

% actual training

DeepGaze II is trained using maximum likelihood learning (see \cite{kuemmerer2015} for an extensive discussion of why log-likelihoods are a meaningful metric for saliency modelling).
If $p(x, y \mid I)$ denotes the probability distribution over $x$ and $y$ predicted by DeepGaze II for an image $I$, the log-likelihood of a dataset is
\[
  \frac{1}{N} \sum_i \log p(x_i, y_i | I_i),
\]
for fixations at locations $(x_i, y_i)$ in the image $I_i$.
This loss function depends on the parameters of the readout network and the kernel size of the Gaussian used to regularize the prediction
(note that it also depends on the parameters of VGG, but we do not retrain them).
As it also is differentiable in these parameters, of-the-shelf optimization techniques can be used to optimize the loss.
Here we use the \textit{Sum-of-Functions-Optimizer} (SFO, \cite{sohldickstein2013}), a mini-batch-based version of L-BFGS.
The full training procedure consists of multiple phases and is visualized in \ref{fig:crossvalidation}.

In the pretraining phase, the readout network is initialized with random weights and trained on the SALICON dataset \parencite{Jiang2015}.
This dataset consists of 10000 images with pseudofixations from a mouse-contingent task and has proven to be very useful for pretraining saliency models \parencite{Huang_2015_ICCV, Kruthiventi2015, jetley2016}.
All images are downsampled by a factor of two.
We use 100 images per mini-batch for the SFO.

The MIT1003 dataset is used to determine when to stop the training process.
After each iteration over the whole dataset (one epoch) we calculate the performance of the model on the MIT1003 (test) dataset.
We wish to stop training when the test performance starts to decrease (due to overfitting).
We determine this point by comparing the performance from the last three epochs to the performance five epochs before those.
Training runs for at least 20 epochs, and is terminated if all three of the last epochs show decreased performance or if 800 epochs are reached.
As it is more expensive to use images of many different sizes, we resized all images from the MIT1003 dataset to either a size of $1024\times 768$ or $768 \times 1024$ depending on their aspect ratio, before downsampling by a factor of two.

After pretraining, the model is fine-tuned on the MIT1003 dataset.
As DeepGaze I showed that overfitting to images is in fact a much larger problem than overfitting to subjects, DeepGaze II is crossvalidated over images:
the images from the dataset are randomly split into 10 parts of equal size.
Then ten models are trained starting from the result of the pretraining, each one using 9 of the ten parts for training and the remaining part for the stopping criterion (following the stopping criteria as above).
We use 10 images per mini-batch in the SFO.

% this paragraph might be redundant with results?
When evaluating on any dataset but the MIT1003 dataset, we use a mixture of these ten models.
This holds specifically for the MIT300 dataset from the MIT Saliency Benchmark.
When evaluating on the MIT1003 dataset for our in-house analyses, for each image we use the model which has not been trained using this image.

\section{Results}

How well does the DeepGaze II model perform on saliency prediction relative to other saliency models?
We first consider this from the standpoint of information theory (information gain explained) evaluated on a subset of the MIT1003 dataset (as used in \textcite{kuemmerer2015, kuemmerer2015deepgaze}), and second present results from the MIT saliency benchmark website on the held-out MIT300 set.

\subsection{Information gain explained}

In \cite{kuemmerer2015}, we described the calculation of information gain explained (an intuitive transformation of log-likelihood).
Information gain tells us what the model knows about the data beyond the baseline model, which here is the image-independent centre bias, expressed in bits / fixation:
\[
  IG(\hat p \| p_\mathrm{baseline}) = \sum_i \log \hat p(x_i, y_i | I_i) - \log p_\mathrm{baseline}(x_i, y_i) 
\]
where $\hat p(x, y | I)$ is the density of the model at location $(x, y)$ when viewing image $I$, and $p_\mathrm{baseline}$ is the density of the baseline model.
Information gain explained relates the model's information gain to the gold standard (crossvalidated prediction of all subjects from all other subjects---sometimes called the ``empirical saliency map'') information gain. 
It is the proportion of the gold standard information gain accounted for by the model.
\[
  \frac{IG(p \| p_\mathrm{baseline})}{IG(p_\mathrm{gold} \| p_\mathrm{baseline})}
\]
where $p_\mathrm{gold}$ is the density of the gold standard model.

To remain consistent with our previously published work \parencite{kuemmerer2015, kuemmerer2015deepgaze}, we evaluate DeepGaze II on a subset of the MIT1003 dataset consisting of all images of size $1024 \times 786$ ($N = 463$).
For each image in this set, there is exactly one model from the fine-tuning crossvalidation procedure that did not use that image for training.
We use the density from this model for evaluation.
This means we are reporting test performance, crossvalidated over images, as opposed to training performance.

The gold standard model is essentially a Gaussian kernel density estimate that predicts one subject's fixations for a given image from the fixations of all other subjects.
That is, the gold standard model is an image-specific prediction crossvalidated over subjects, and as for the models we report test not training performance.

\begin{figure}
	\centering
	\includegraphics[width=.7\textwidth]{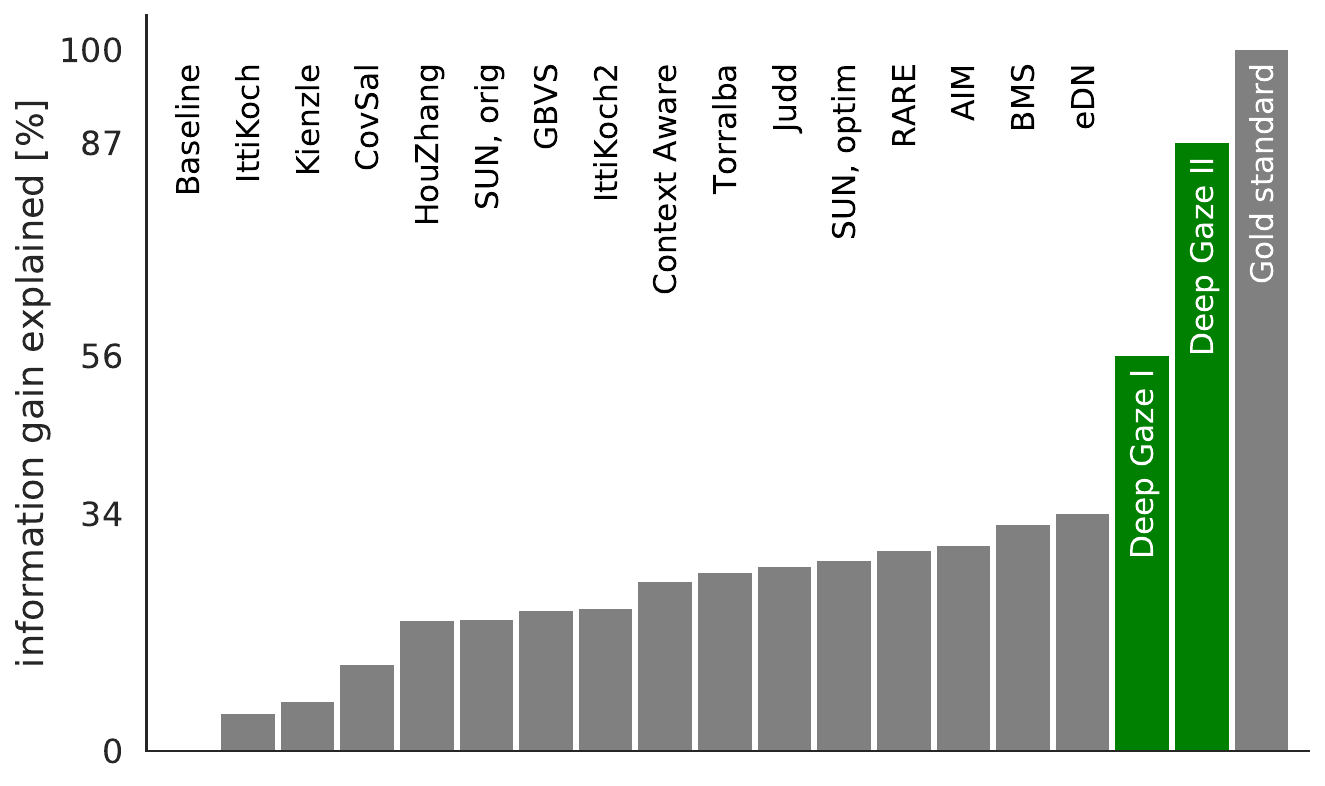}
	\caption{Model performance (information gain explained as a percentage of the gold standard model's information gain relative to the baseline model) for a selection of models from the MIT Benchmark, DeepGaze I and DeepGaze II. 
	The eDN model (state-of-the-art in 2014) explained 34\% of the explainable information gain, and DeepGaze I explains 56\%.
	DeepGaze II gains a substantial improvement over DeepGaze I, explaining 87\% of the explainable information in the evaluation set.
	}
    \label{fig:performance_barplot}
\end{figure}

Figure \ref{fig:performance_barplot} shows the information gain explained for DeepGaze II against that for DeepGaze I and the models evaluated in \cite{kuemmerer2015}.
DeepGaze II accounts for 87\% of the explainable information gain, a substantial improvement from DeepGaze I's 56\%, and begins to approach the upper limit (according to the gold standard) of performance in saliency prediction.
Note that we currently do not include models that improved on DeepGaze I on the MIT benchmark (SALICON, DeepFix and PDP) in this evaluation because the code for these models is not publically available.

\begin{figure}
  \begin{center}
     \includegraphics[width=0.45\textwidth]{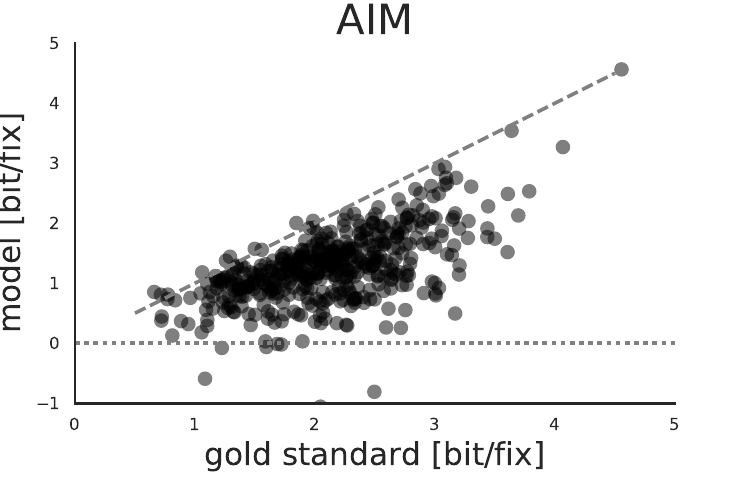}
     \includegraphics[width=0.45\textwidth]{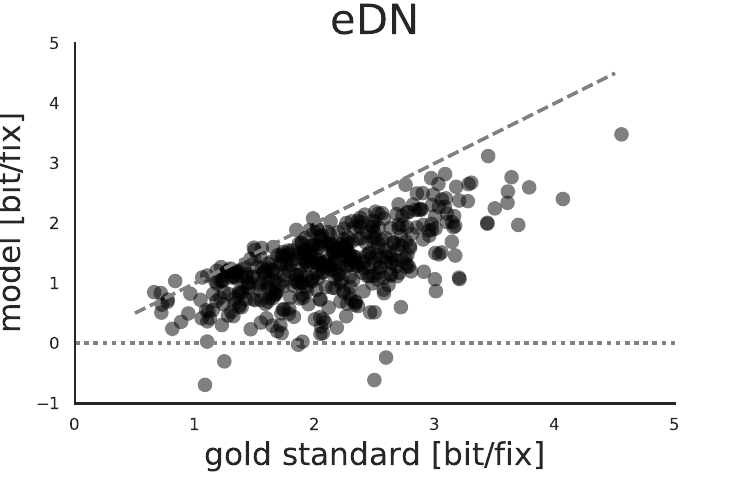}
     \vspace{2ex}
     \includegraphics[width=0.45\textwidth]{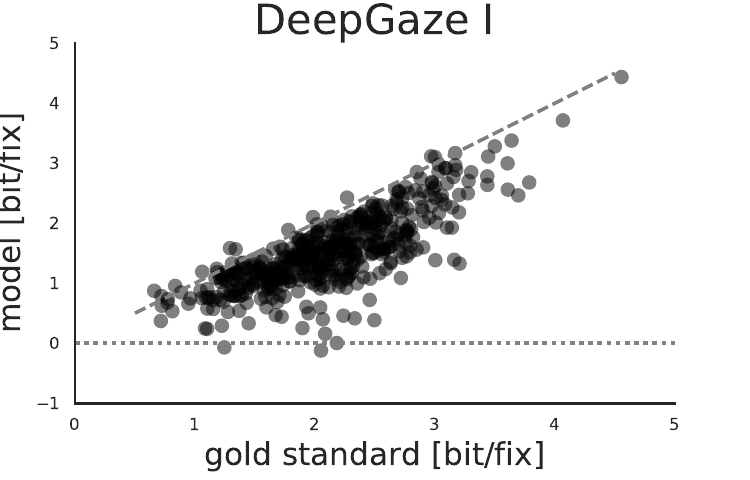}
     \includegraphics[width=0.45\textwidth]{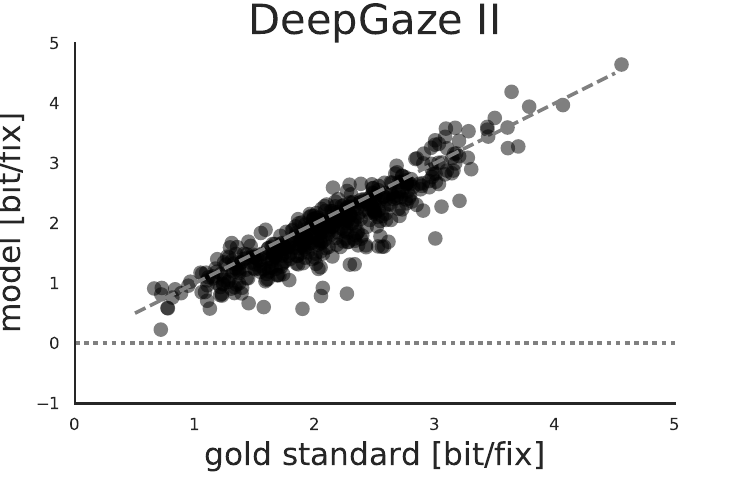}
    \end{center}
    \label{fig:performance_scatterplots}
    \caption{Gold standard information gain against model information gain 
    relative to the baseline model, for AIM, eDN, DeepGaze I and DeepGaze II.
    Each point is an image in the subset of the MIT1003 dataset used for evaluation.
    DeepGaze II is highly correlated with the gold standard, and is the only model for which
    no images show negative information gain (i.e. for which the model's prediction is worse
    than the pure centre bias).}
\end{figure}

We can also evaluate candidate models according to their performance relative to the gold standard for each image in the dataset (Figure \ref{fig:performance_scatterplots}).
Here, one can see that the AIM, eDN and DeepGaze I model predictions fall largely below the gold standard, and all include a number of images with negative information gain (meaning that the models make worse predictions than the baseline for those images).
DeepGaze II clusters much closer to the gold standard predictions (diagonal line) and there are no images for which its prediction is worse than the baseline.
Note that it is possible to have images for which the model prediction is \textit{better} than the gold standard.
There can be at least two reasons for this: first, it can be that fixations cluster in smaller areas than predicted by the gold standard (recall that the gold standard kernel size is learned over all images); second, there could be subjects who are inconsistent relative to other subjects but still look at areas that a model can predict.
In this case the gold standard model performs poorly when predicting that subject relative to the model (recall that the gold standard performances are test performances). 

\subsection{MIT saliency benchmark}

\begin{table}
  \begin{center}
    \begin{tabular}{p{6cm}|c|c}
           Model & \phantom{x}AUC\phantom{x} & \phantom{x}sAUC \\ \hline
           DeepGaze II & \textbf{88\%} & \textbf{77\%} \\
           SALICON     & 87\% & 74\% \\
           DeepFix     & 87\% & 71\% \\
           DeepGaze I  & 84\% & 66\% \\

    \end{tabular}
  \end{center}
  \label{tab:mitbenchmark}
  \caption{DeepGaze II performance in the MIT Saliency Benchmark. DeepGaze II reaches top performance in both AUC and sAUC.
  Note that we use saliency maps without center bias for the sAUC result (see text for more details).}  
\end{table}

The area under the ROC curve (AUC) metric expects saliency maps to include the centre bias, whereas shuffled AUC expects models to exclude the centre bias \parencite{Barthelme, kuemmerer2015}.
Because the DeepGaze II architecture makes it trivial to include or exclude the centre bias into the model prediction, we submitted two sets of saliency maps to the MIT benchmark: one uses the centre bias trained on the MIT1003 dataset, the other uses a uniform centre bias.
In addition, because the MIT Benchmark requires submission of model predictions as JPEG images, we quantised the log density for each image into 256 values such that each value receives the same number of pixels.

Table \ref{tab:mitbenchmark} reports the results of evaluating DeepGaze II on the MIT saliency benchmark (the held-out MIT 300 set).
DeepGaze II beats the nearest competitors SALICON and DeepFix by one percent.
For shuffled AUC, DeepGaze II beats the nearest competitors by a larger margin (note that this could be due in part to those models not excluding centre biases).

\begin{figure}
  \begin{center}
    \includegraphics[width=\textwidth]{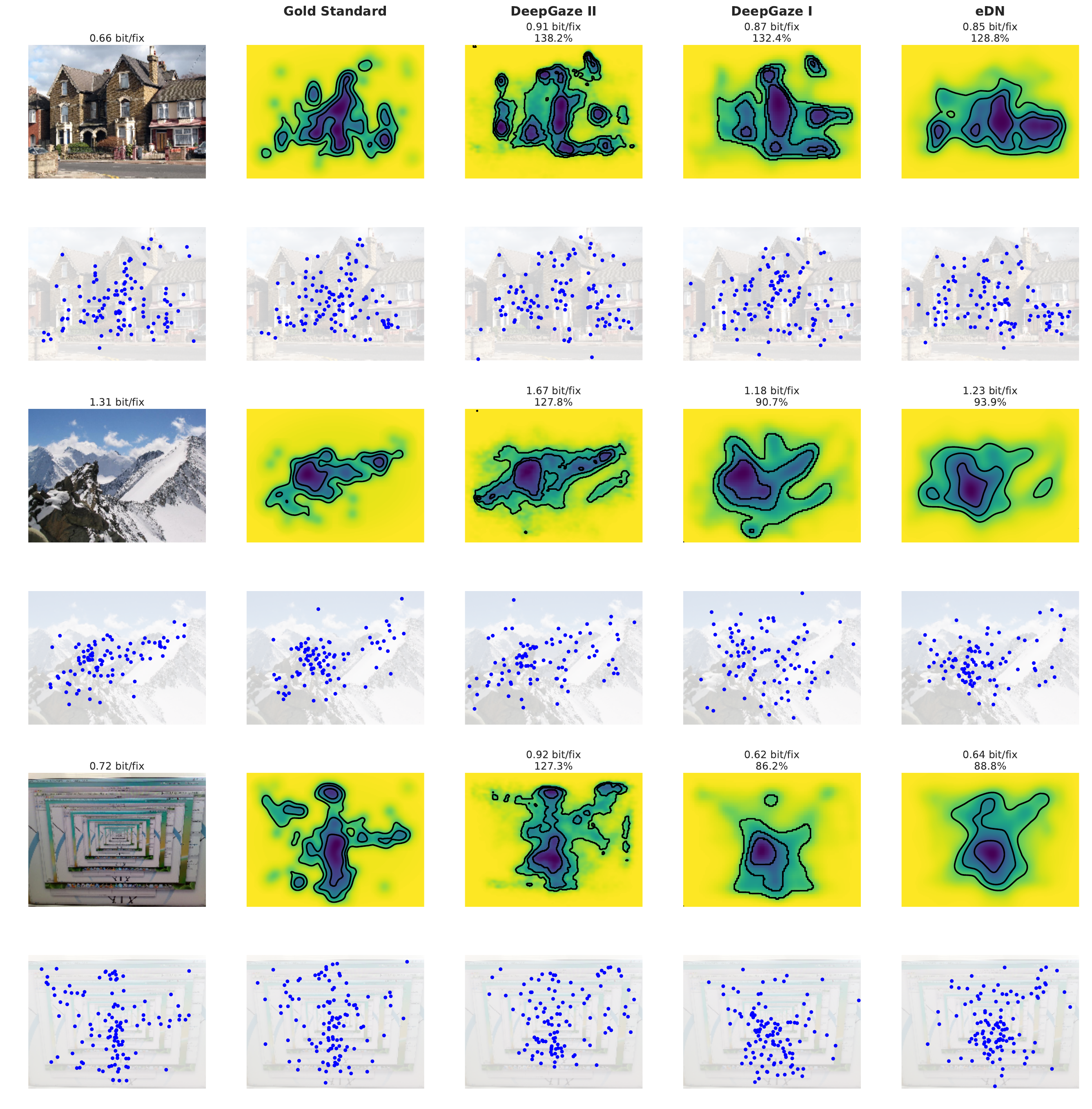}
  \end{center}
  \caption{The three images for which DeepGaze II had the highest information gain explained.
  	For each unique image, the leftmost column shows the image itself (top) and the empirical fixations (bottom).
  	The remaining columns show model predictions for the gold standard model, DeepGaze II, DeepGaze I and the eDN model respectively.
  	The top row visualises probability densities, in which contour lines divide the images into four regions, each of which is expected to receive equal numbers of fixations.  
  	The bottom row shows fixations sampled from the model (see text for details). 
  	Sampled fixations can be compared to the empirical fixations to gain additional insight into model performance.
  }
  \label{fig:by_ige_best}
\end{figure}

\begin{figure}
  \begin{center}
    \includegraphics[width=\textwidth]{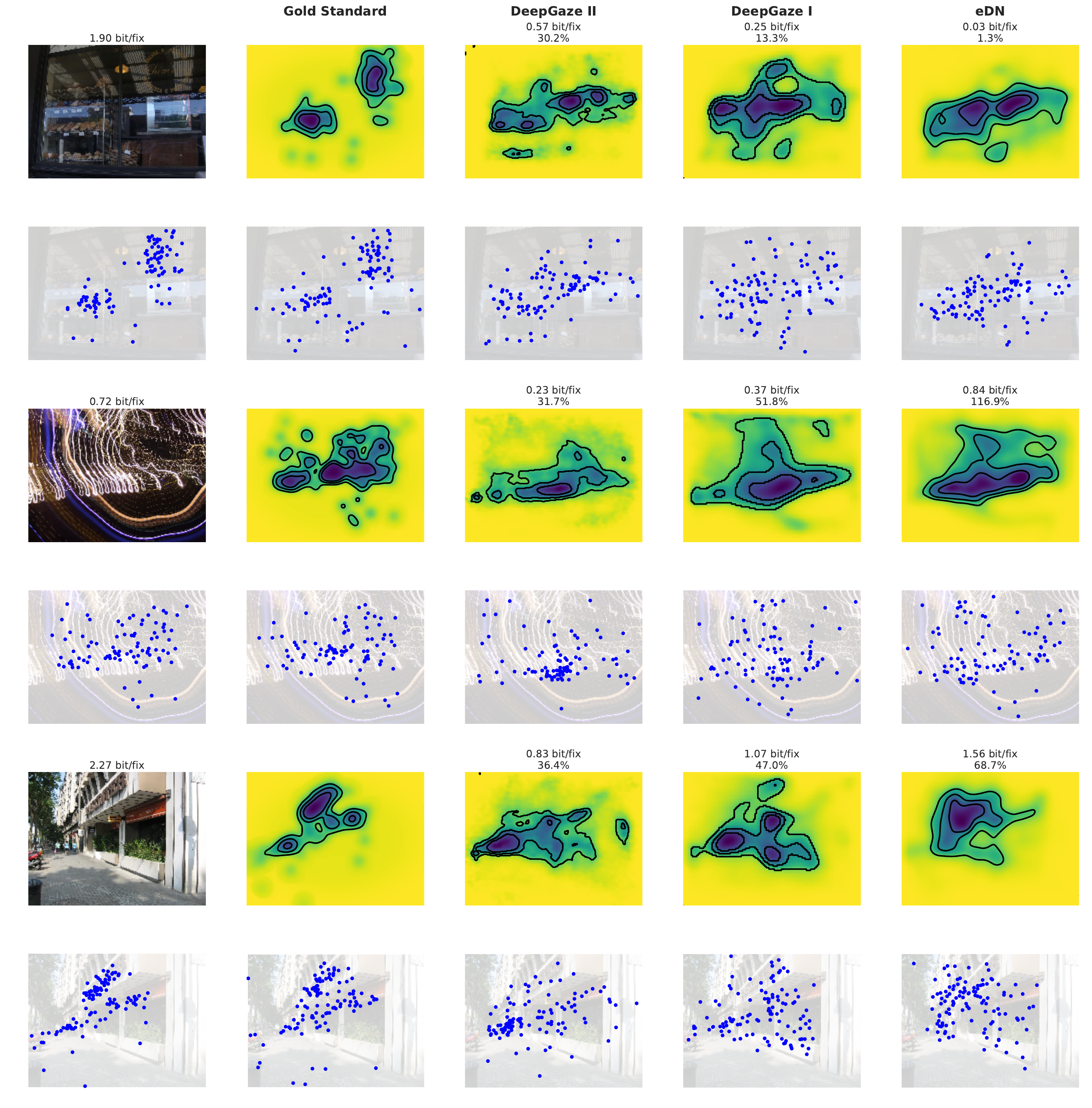}
  \end{center}
  	\caption{The three images for which DeepGaze II had the lowest information gain explained.	
  	}
  \label{fig:by_ige_worst}
\end{figure}

\subsection{Model prediction examples}
% paragraph about specific image examples.
Figure \ref{fig:by_ige_best} shows the three images for which DeepGaze II explained the most of the explainable information gain in the patterns of fixations, and Figure \ref{fig:by_ige_worst} shows the worst.
For visualising probability densities, we include three contour lines which together divide the map into four regions. 
Each region has the same probability mass: that is, the model expects each area to receive the same number of fixations on average.
If the dark areas are very concentrated, then the model expects a small area to receive most of the fixations.
In addition, for each image we sample from each model to obtain the same number of fixations as for the ground truth fixations.
Sampling is straightforward because the density predicted by the model is a multinomial distribution over the pixels.
This allows an intuitive comparison of model and data.
Note that both of these analysis approaches are only possible using a probabilistic model.

Some interesting patterns to consider include the first image in Figure \ref{fig:by_ige_worst}, 
which is a photograph of a bakery shopfront.
Humans fixate on the baked goods (which DeepGaze II captures) and on the store logo imprinted on the window
in the upper right of the image (which DeepGaze II fails to capture, presumably because it does not detect the
low-contrast, partially-occluded text). 
For the third image of Figure \ref{fig:by_ige_worst}, people fixate on the signage above the storefront, which in the image is distorted by perspective projection.
Both DeepGaze I and II appear to miss this text.
This might be because both the VGG and AlexNet fail to provide features sensitive to such distorted text, or because distorted text is so rare in the training set that the contributions of these features are downweighted by our training procedure.
In either case, these two examples highlight one potential avenue for model improvement (better training on text).

\subsection{Reasons for improvement over DeepGaze I}

\begin{figure}
	\centering
	\includegraphics[width=.7\textwidth]{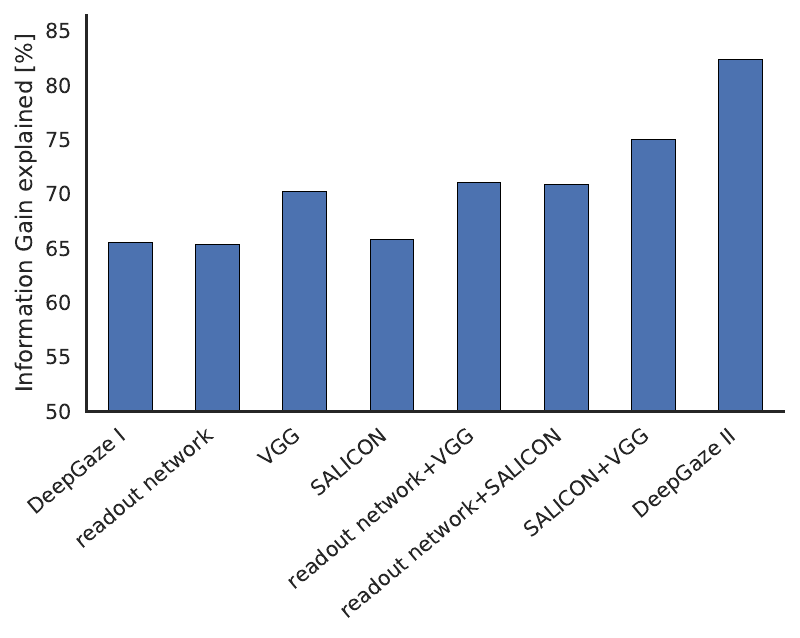}
	\caption{The contribution of the primary changes from DeepGaze I to DeepGaze II in terms of improving information gain explained.
	The most important contributions to the improvement are using VGG features and pre-training on the SALICON dataset.
	}
    \label{fig:performance_contributions}
\end{figure}

Why is DeepGaze II better than DeepGaze I?
We quantified the contributions of the three primary changes from DeepGaze I to DeepGaze II on the MIT1003 dataset\footnote{
Note that this is not the original DeepGaze I model as presented in \cite{kuemmerer2015}. 
Here we have trained on the full MIT1003 dataset and used the same scheme of crossvalidation over images as described in this paper.
}.
As seen in Figure \ref{fig:performance_contributions}, the largest single improvement is brought by using the pretrained VGG features in place of AlexNet (though we also include more channels from VGG than from AlexNet).
Using the readout network rather than a linear regression slightly decreases performance when considered independently, likely due to overfitting.
Training on the SALICON dataset marginally improved performance.
Combining SALICON pretraining with the VGG features yields the largest intermediate model performance improvement.

\begin{figure}
  \begin{center}
    \includegraphics[width=\textwidth]{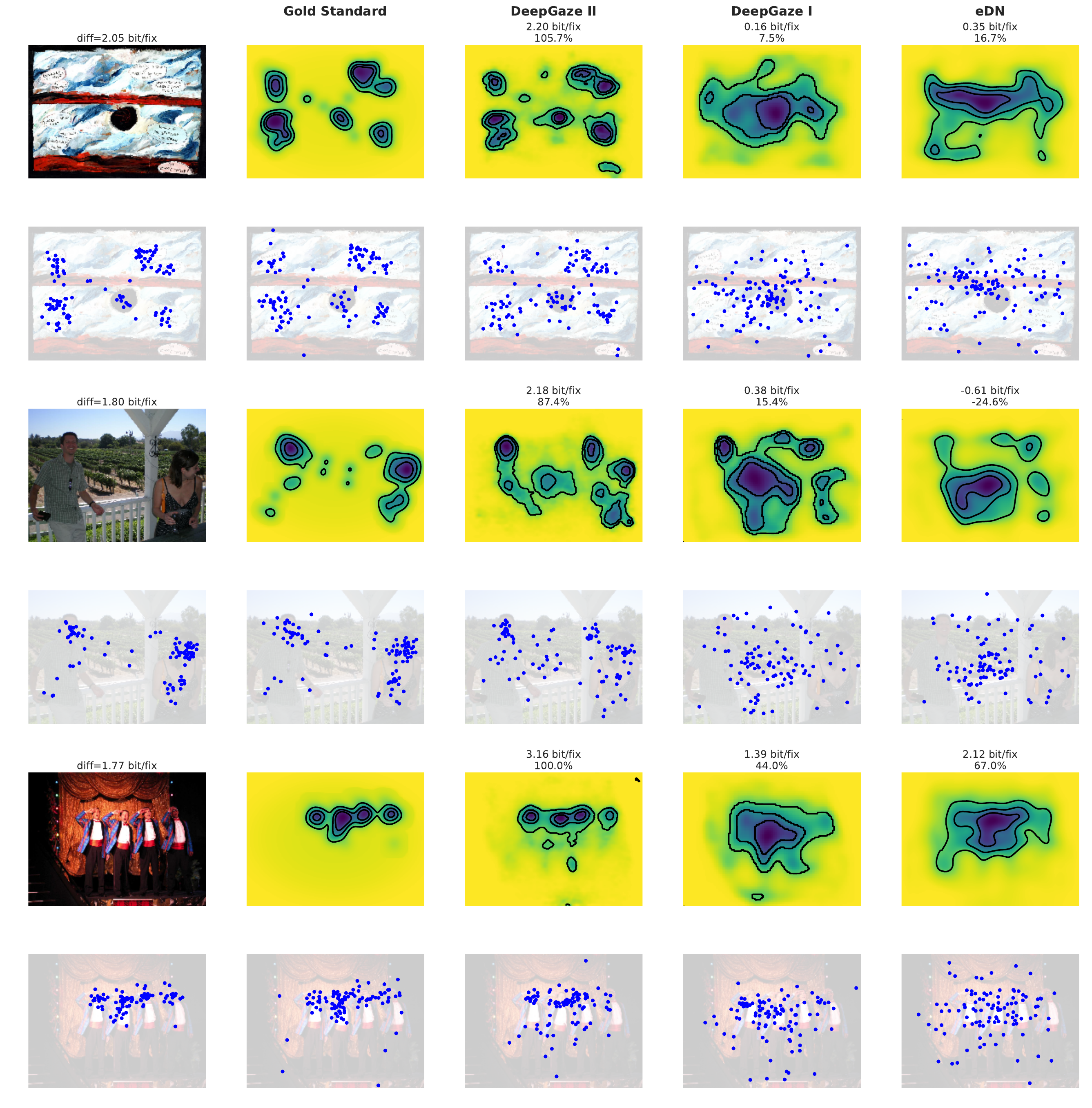}
  \end{center}
  \caption{The three images for which DeepGaze II most improves predictions relative to DeepGaze I in terms of information gain.}
  \label{fig:by_ig_diff_dg1_best}
\end{figure}

\begin{figure}
  \begin{center}
    \includegraphics[width=\textwidth]{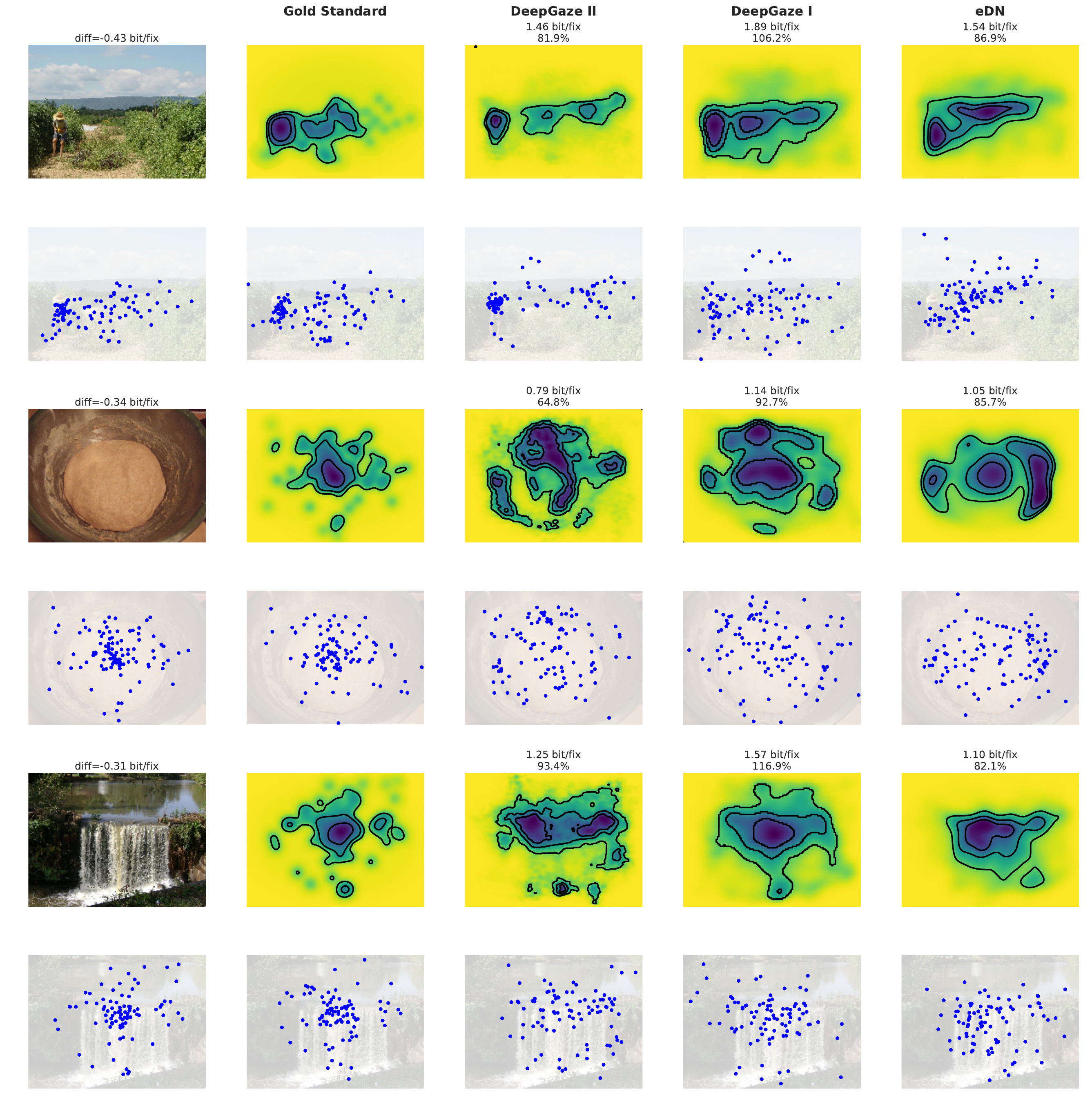}
  \end{center}
  \caption{The three images for which DeepGaze II most fails in prediction relative to DeepGaze I.}  
  \label{fig:by_ig_diff_dg1_worst}
\end{figure}

We additionally provide examples of images for which DeepGaze II improves most from DeepGaze I (Figure \ref{fig:by_ig_diff_dg1_best}) and performs worse than DeepGaze I (Figure \ref{fig:by_ig_diff_dg1_worst}) in terms of information gain differences (in bit/fix).
The improvement for the first image in Figure \ref{fig:by_ig_diff_dg1_best} seems to be driven by better recognition of text, whereas for the second and third images DeepGaze II seems to benefit from improved (or more spatially-specific) face and person detection.

\section{Discussion}

Here we have presented DeepGaze II, a model of saliency prediction that uses transfer learning from the VGG-19 network to achieve state-of-the-art performance.
Information gain explained is able to quantify precise differences between models, and shows the clear improvement gained by DeepGaze II (note however, that some high-performing models were not included in these evaluations because their code is not publically available).
Our model is also ranked first on the held-out MIT300 benchmark according to AUC and shuffled AUC, the most commonly-reported evaluation metrics.
Note however that here, at least for AUC, the difference between DeepGaze II and other models is modest.

% Why does DG do better relative to other deep learning methods?
Why does DeepGaze II perform better relative to other models that also use deep features?
We believe this could be because, at least in part, we do not retrain the VGG features.
While this reduces the model space, it also greatly reduces the number of parameters that must be learned from data, reducing the chance of overfitting.
Furthermore, since we only use $1 \times 1$ convolutions on top of this, we cannot learn new features that are substantially different from VGG: only a pointwise nonlinearity is possible.
These two aspects of our model therefore represent a much more stringent test of the transfer success of deep features.
%The fact that they work so well suggests that VGG features return quite precise location information despite being trained only on object detection and having large receptive fields.

We provide a web service to calculate DeepGaze II predictions for arbitrary images at \url{http://deepgaze.bethgelab.org}.

\printbibliography

\end{document}